# LPM: Learnable Pooling Module for Efficient Full-Face Gaze Estimation

Reo Ogusu and Takao Yamanaka

Department of Information and Communication Sciences, Sophia University, Tokyo, Japan

*Abstract*—Gaze tracking is an important technology in many domains. Techniques such as Convolutional Neural Networks (CNNs) have allowed the invention of the gaze tracking method that relies only on commodity hardware such as the camera on a personal computer. It has been shown that the full-face region for gaze estimation can provide a better performance than the one obtained from eye image alone. However, a problem with using the full-face image is the heavy computation due to the larger image size. This study tackles this problem through compression of the input full-face image by removing redundant information using a novel learnable pooling module. The module can be trained end-to-end by backpropagation to learn the size of the grid in the pooling filter. The learnable pooling module keeps the resolution of valuable regions high and vice versa. This proposed method preserved the gaze estimation accuracy at a certain level when the image was reduced to a smaller size.

## I. INTRODUCTION

Many researches have focused on estimating the human gaze for its importance in areas such as human-computer interaction [1–3], bio-medical engineering [4], and autonomous driving [5, 6]. Gaze tracking techniques have evolved over time, though most of the early methods require expensive and invasive hardware. Furthermore, these solutions are heavily restricted in lighting conditions or in head poses [7]. While these factors had prevented the technology to be applied in daily life, the latest development in Convolutional Neural Networks (CNNs) allowed the invention of gaze tracking techniques which rely on commodity hardware such as the camera on a personal computer [8] or a mobile phone [9]. Although these images are diverse in lighting conditions, head poses, and appearances, CNNs has achieved good results.

The work done by Krafka et al. [9] has shown that the CNN architecture with the input from an eye image, a face image, and a face grid can greatly improve the performance of previous architectures with the input from an eye image alone. Zhang et al. [10] has taken this idea further and has proposed an appearance-based method that takes the full-face image as the input. The CNN architecture has used the spatial weights mechanism to enhance important information across different regions of the full-face. Although the method paved the way for the gaze tracking technology to withstand everyday settings, the use of a full-face input is inefficient regarding the computational cost and memory usage. On the contrary, the use of a smaller resized face image can reduce the computational cost, though the estimation accuracy is low. The goal of this work is to find a method that can resize the full-face image in the way that the accuracy is preserved.

To achieve this goal, we propose the learnable pooling module (LPM) that efficiently downsamples a full-face image and can be trained in an end-to-end fashion. It is known that regions other than eyes are contributing to the estimation but the degree differs among parts [10]. In the proposed method, it is shown that after sufficient training of the network, the pooling layer can compress an input face image in a way that important regions are kept at high resolutions. This method of using the LPM for compression of the full-face allows better performance on gaze-estimation accuracy than resizing the image as it is using the standard average pooling. Contributions of this paper are as follows:

- To develop a novel learnable pooling module which can be trained in an end-to-end fashion.

- To propose a method that can efficiently compress a face image using the novel pooling module for gaze estimation.

- To reduce the computational cost for inference while retaining the estimation accuracy.

## II. RELATED WORKS

The work in this paper is related to the previous works on 2D gaze estimation using appearance-based methods. Moreover, the study takes the idea from region of interest (ROI) pooling and image compression using saliency maps for the development of the LPM.

### A. Gaze Estimation

The appearance-based gaze estimation method is defined as the regression task that directly maps an eye image to the gaze direction. For the appearance-based method, the calibration step to collect person-specific training data is known to be helpful for the estimation task. Although the calibration can improve the performance [11], it limits the estimator's capacity to be generalized. However, it has been shown that the data-driven approach is effective for learning a gaze estimator that performs well without calibration [7] and robust to different head poses [8]. For this reason, many works have focused on creating labeled training datasets specifically designed for gaze estimation in non-restricted settings [8, 12, 13] and some at a large scale [9]. Using these datasets for training has allowed for the gaze estimator to be performed better in various lighting conditions, head poses, and facial expressions, even without calibration.

Furthermore, the appearance-based gaze estimation methods can be categorized into either the 3D or 2D gaze estimation depending on the target [10]. The 2D gaze estimation is simpler in concept than the 3D models. The aim of the 2D gaze estimation is to output the gaze location coordinate on the target screen, while the 3D gaze estimation is to output 3D coordinate. The 2D gaze estimation task can be expressed as a regression from an input image $I$ to the 2D gaze location $p$ in the coordinate system of the target screen, as $p = f(I)$ where $f$ is the regression function. The core idea in the 2D gaze estimation is to learn $f$. The work in this paper focused on the 2D gaze estimation.

Recent works have proven that CNNs are effective for realizing the 2D gaze estimator that works in daily life settings. Zhang et al. [8] have proposed a CNN-based method where the model takes two inputs of a 2D head angle and a low-resolution eye image (of size $36 \times 60$). Krafka et al. [9] have established the more data-driven approach of an end-to-end CNN for gaze estimation. By relying only on cropped images of the face and the eyes, accompanied by the corresponding face grid, the network outperformed the existing methods by a large margin. Interestingly, the model achieved the performance close to the state-of-the-art (SOTA) only using the face and the face grid as the input. Taken this result, a full-face appearance-based gaze estimator where a full-face image (of size 448×448) is fed into a CNN with the spatial weights mechanism has been proposed, leading to the SOTA performance [10].

These recent findings indicate the importance of facial regions other than eyes, though the degree of importance is not yet known. Moreover, the large size input is not practical especially for the full-face appearance-based gaze estimation where a large-scale dataset is essential for training the estimator that is robust beyond lab settings.

*B. Downsampling*

In a neural network, it is common to insert a pooling layer to reduce the spatial size of the feature map for realizing translational invariance. Usually, a pooling layer uses a grid of a fixed size for sampling. On the contrary, ROI pooling [14] which converts an input rectangular region of arbitrary size into fixed-size features has been proposed. Given the input feature map and an ROI of size $w \times h$ at an arbitrary position, ROI pooling divides the ROI into $K \times K$ ($K$ is a free parameter) cells and outputs a $K \times K$ feature map. The ROI pooling has been further advanced to the Position-Sensitive (PS) ROI Pooling as proposed in [15] to fasten object detection tasks.

The proposed method is also related to image compression using saliency maps, which compresses the image so that the regions with high visual saliency are reserved with higher resolution than the regions with low saliency [16]. Following the idea of the pooling layer with arbitrarily sized grid and the compression method that adjusts the resolution of certain regions, we propose a novel pooling module that can learn the grid size of the pooling filter.

III. METHOD

*A. Learnable Pooling Module (LPM)*

One strategy to reduce the computational cost of CNNs is to reduce the size of the input image. Although the accuracy drops as the image size decreases, the proposed method of the LPM compresses an image with the minimal accuracy drop.

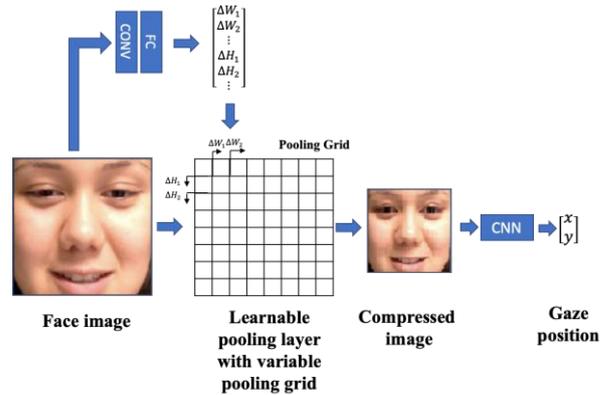

*Fig. 1 Overview of compressing face image with the LPM. The module takes input image and pass the image to two layers. The output of the layers are the offsets to be used to move the border positions and adjust the size of the pooling grid.*

This module operates on the 2D domain with the same operation applied across channels. The module is described in 2D in this section for clarification.

Fig. 1 illustrates the overview of the learnable pooling architecture. The network processes a given input image through a convolutional layer (CONV) and a fully connected layer (FC), and outputs the offsets for border positions of the pooling grid with different pooling sizes depending on its positions. The pooling grid is applied to the input image at the "learnable pooling layer" for the efficient image compression. The average value is computed for each grid of the LPM. The pooling grid derived from the border offsets is learned from training data in the end-to-end fashion and is unique to each face image input to the network.

*B. Forward Path*

The network processes the given input image through two layers. The first is a normal convolutional layer with a kernel size of $3 \times 3$ and a stride of 2. It is followed by a fully connected layer. This layer outputs the offsets $\Delta w$ for border positions of a pooling grid. Each positional parameter is a real number. A mechanism for limiting the offset values is applied to assure that the orders of the borders are not inverted. When the offset value is too large for the updated position to overpass the adjacent borders, it is adjusted to be one pixel more or less than the adjacent border to maintain the order.

Using the border positions calculated with the offsets from the two layers, the learnable pooling layer updates the position of each border of the pooling filter. Given the input image $x$ of size $w \times h$, the layer divides the image into $K \times K$ cells in the grid using the border positions and outputs feature map $y$ of size $K \times K$. For the $(i,j)$-th cell in the grid ($0 \leq i, j < K$) the output is represented as follows.

$$y(i,j) = \frac{1}{n_{ij}} \sum_{\boldsymbol{p} \in grid(i,j)} x(\boldsymbol{p}) \qquad (1)$$

$n_{ij}$ is the number of pixels in the cell.

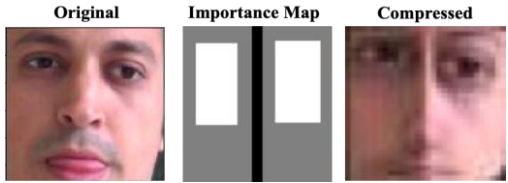

*Fig. 2 Overview of compressing face image with importance map. The compression method takes an original image (left) and a manually set importance map (center) as input and outputs a compressed image (right).*

As the network is trained, the region contributing to the gaze tracking task is averaged with a high resolution (filtered with smaller pooling cells) and the non-contributing region with a low resolution (filtered with larger pooling cells).

*C. Backward Path*

Since the parameters of the offsets in the LPM are not analytically differentiable, the gradients are numerically obtained. The gradient for the border $j$ of the column of the image can be computed as

$$\frac{\partial y(i,j)}{\partial p_j} \cong \frac{y(i,j)' - y(i,j)}{h} \quad (2)$$

where $y(i,j)'$ is the $(i,j)$-th element of the output feature map $y$ after the position of the border $j$, $p_j$, was moved with the displacement of $h$. The displacement of h was set to 1 in the experiment. When the border position calculated with the offset in the forward path overpasses the adjacent border, the gradient for the offset is set to 0, because the offset is limited in the forward path. The gradients for the borders of the rows can be computed in the same way.

IV. EXPERIMENT SETUP

*A. Preliminary Experiment*

Prior to the main experiment, a preliminary experiment was conducted to validate whether the preservation of the resolution on important regions has an effect on the gaze estimation task. Our hypothesis here is that the regions other than the eyes are contributing to the estimation but the degree differs among parts; for instance, the eyes being the most important and the mouth being the least. A downsampling algorithm that adjusts the grid size of the pooling filter with the manually designed importance map was developed for the experiment. The importance map put weights on important regions on the face image. Given the importance map corresponding to the face image, it is downsampled into the small-size. The white areas of the map in Fig. 2 are given importance of 1. The inner peripheral importance value is greater than 0 but smaller than 1 to put weight on gray areas of the importance map. The remaining regions are given a value of 0. The algorithm compresses an image by adjusting the grid size of the pooling filter depending on the importance values given to the pixels [17]. Finally, the compressed image is obtained by averaging the pixel values of the original face image within the range of pixels in a cell of the grid. The resolution is preserved for areas with high importance and vice versa. Importance maps were designed so that more pixels are allocated for the eyes which are assumed to contain valuable information. The maps were created using the eye coordinates provided by the dataset. Various image sizes and importance values were tested. The experiment was conducted by training SqueezeNet [18] using the images compressed with the importance map. The full-face images compressed with the standard average pooling was also evaluated for comparison.

*B. Experiment for Learnable Pooling Module*

Although the importance maps were manually designed in the preliminary experiment, the proposed LPM can learn the pooling grids from the data by training. Through the implementation of the LPM, it was realized that the full-face image was reduced to a smaller size in a way that the accuracy was preserved, with higher resolution on the important regions.

The experiments were conducted by comparing the results between images downsampled with the LPM and the standard average pooling. Although the original size of a full-face image was $448 \times 448$ [10], a full-face image used for the experiment was resized to a smaller size in order to avoid heavy computation. Concretely, a full-face image of size $112 \times 112$ was prepared and downsampled to the size $30 \times 30$.

*C. Data Preparation*

The experiments were conducted on the MPIIGaze dataset [8], the same dataset Zhang et al. used in their paper [10]. Since the number of samples for every 15 subjects ranged from 1,498 to 34,745, the same image selection process in the paper [10] was performed, where 3000 unique images were prepared for each participant. Although the proposed method relies on the face detector for cropping the face, the extraction of the face was already completed in the MPIIGaze dataset.

*D. Implementation Details*

The CNN model used for the experiments is Squeeze Net proposed in [18]. Although the conventional method [10] which achieved the state-of-the-art for the appearance-based gaze estimation has used AlexNet [19], SqueezeNet was chosen for the base architecture of the CNN component in Fig.1, because SqueezeNet is a model that has achieved image-classification performance similar to AlexNet with 50 times fewer parameters [18]. Moreover, the accuracy is expected to be improved since the size of the feature map is kept larger within the model compared to AlexNet. A face grid, a binary mask that indicates the location and the size of the face in the original frame, was used as an additional input, as in [10]. The face grid of the size $50 \times 50$ was passed through two fully connected layers, whose output was concatenated with the output from the second last layer of the CNN network.

The learning rate for the fully connected layer that outputs the offsets of the borders is especially important for the LPM to be trained properly. If the learning rate is too large, the updated border position overpasses the neighboring border position. Because the gradient is returned as 0 in that case, the training does not proceed. To avoid this, the learning rate for the fully connected layer in Fig. 1 was reduced by a factor of $10^{-1}$ when more than 20 % of the total number of borders overpassed their neighboring borders (dynamic learning rate). Additionally, the rate was increased by 10 when less than 10% of the total number of adjustable borders overpassed the neighboring borders. The initial learning rate was set to 0.01 and was adjusted every 10 iterations during the training. The learning rate was limited in the range of $10^{-6} \leq lr \leq 10^{-1}$.

TABLE I. PERFORMANCE OF GAZE ESTIMATION USING FULL-FACE IMAGE (448 × 448) COMPRESSED WITH IMPORTANCE MAPS.

| Image Size (pixels) | Downsampling Method | Min Error (mm) | Estimation time [ms/image] |
|---|---|---|---|
| 30 × 30 | Average pooling | 72.580 | 11.814 |
|  | Importance Map | 62.255 |  |
| 112 × 112 | Average pooling | 48.418 | 22.089 |
|  | Importance Map | 42.218 |  |
| 448 × 448 | None [10] | 42.0 | 220.320 |

The model was implemented using Chainer [20]. The CNN component of the model was pre-trained with ImageNet [19]. The loss function was the Euclidean distance between the predicted and the ground-truth gaze coordinates in the target screen. SGD was used for the optimizer. The batch size was set to 64.

*E. Evaluation*

Similar to [10], the accuracy is defined as the average of the Euclidean distances from ground-truth coordinates to the estimated gaze coordinates. The dataset contained 15 subjects, where leave-one-person-out cross-validation was conducted. The training was stopped at the iteration 32200.

## V. EXPERIMENTAL RESULTS

*A. Preliminary Results*

The results for the preliminary experiments are shown in Table I. It can be seen from the table that the efficiently compressed images with the manually designed importance maps had a better accuracy than the image downsampled with the standard average pooling. Furthermore, the last row of Table I is the state-of-the-art result from [10] using the full-face image of size 448 × 448. Our result with the smaller size of 112 × 112 had a similar accuracy with much faster estimation time. This result showed that the preservation of resolution in important regions enables better accuracy on the task. Figure 2 shows the importance map with the highest accuracy. Although an improvement has been observed, this method to preserve resolution is not necessarily the best since the maps were designed by hand. Therefore, the LPM was developed so that the network can learn the regions to preserve the resolution by training from data.

*B. Results for Learnable Pooling Module*

The results with the LPM, compared with the results using the standard average pooling and the manually designed importance map, are shown in TABLE II. The network with the LPM had a better accuracy than the network which used the average pooling for downsampling. The difference was more significant for the LPM with dynamic learning rate (statistically significant by the paired t-test: $2.075 \times 10^{-3}$).

TABLE II. PERFORMANCE OF LPM. IMAGE OF SIZE 112 × 112 WAS COMPRESSED TO SIZE 30 × 30.

|  | Downsampling Method | Min Error (mm) |
|---|---|---|
| Conventional | Average pooling | 72.580 |
| Proposed (manual) | Importance Map | 63.065 |
| Proposed (learned) | LPM with static learning rate (0.01) | 70.692 |
|  | LPM with dynamic learning rate | 65.801 |

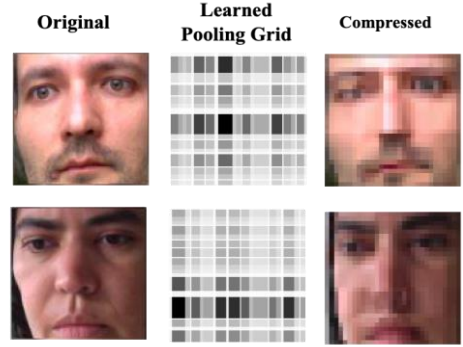

*Fig. 3 Examples of the output from the LPM after training. Left is the original image of size 112 × 112 and middle is the visualization of learned pooling grid. Right is the output image from the module.*

Examples of the compressed images with the LPM after training are shown in Fig. 3 (right). The corresponding visualization of the pooling grid of LPM for each image is shown in the middle column. The pixel with a smaller sampling cell is shown in white, while the pixel with a larger sampling cell is shown in black. It can be seen from the figure that the LPM preserves the resolution of the eyes.

Although the image downsampled with the LPM performed better than the image downsampled with the standard average pooling, the accuracy was slightly higher for the full-face image compressed manually with the importance map. This result may be due to the full-face image output of the LPM having lower resolution in the eye resolution compared to the image compressed with the importance map. Since the dynamic learning rate greatly improved the accuracy greatly, the optimization method for the training needs to be studied to learn better pooling grid for LPM.

## VI. CONCLUSION

A novel learnable pooling module which can be trained in an end-to-end fashion was introduced in this study. The results showed that by using the LPM a face image was efficiently compressed in a way that the accuracy was preserved, resulted in reducing the computational cost for inference. Although further tuning to the module is still needed, the results on the compressed images are promising. Since the LPM is the method which preserves resolution on regions of high importance, the proposed method can be applied broadly in the field of image analysis.